\documentclass[runningheads]{llncs}
\usepackage[T1]{fontenc}
\usepackage{graphicx}
\usepackage{algorithm}
\usepackage{algorithmic}
\usepackage{hyperref}
\usepackage{xcolor}
\usepackage{enumitem}
\usepackage{amsmath,amsfonts,bm}
\usepackage{multirow}
\usepackage{multicol}

\usepackage[misc]{ifsym}

\usepackage{xcolor}
\usepackage{graphicx}
\usepackage{subfig}

\begin{document}
\title{Personalized Education with Ranking Alignment Recommendation}

\author{
Haipeng Liu\inst{1} \and
Yuxuan Liu \inst{2} \and
Ting Long \Letter \inst{3}
}

\institute{School of Artificial Intelligence, Jilin University \\ \email{liuhp22@mails.jlu.edu.cn}\\
\and
School of Artificial Intelligence, Jilin University \\
\email{liuyuxuan23@mails.jlu.edu.cn}\\
\and
School of Artificial Intelligence, Jilin University\\
\email{longting@jlu.edu.cn}}

\maketitle   

\pagestyle{empty}

\begin{abstract}
Personalized question recommendation aims to guide individual students through questions to enhance their mastery of learning targets. Most previous methods model this task as a Markov Decision Process and use reinforcement learning to solve, but they struggle with efficient exploration, failing to identify the best questions for each student during training. To address this, we propose Ranking Alignment Recommendation (RAR), which incorporates collaborative ideas into the exploration mechanism, enabling more efficient exploration within limited training episodes. Experiments show that RAR effectively improves recommendation performance, and our framework can be applied to any RL-based question recommender. Our code is available in \url{https://github.com/wuming29/RAR.git}.

\keywords{question recommendation  \and online education \and reinforcement learning \and ranking.}
\end{abstract}

\section{Introduction}
\label{sec:intro}
\begin{figure}[t]
\centering
\includegraphics[width=0.8\columnwidth]{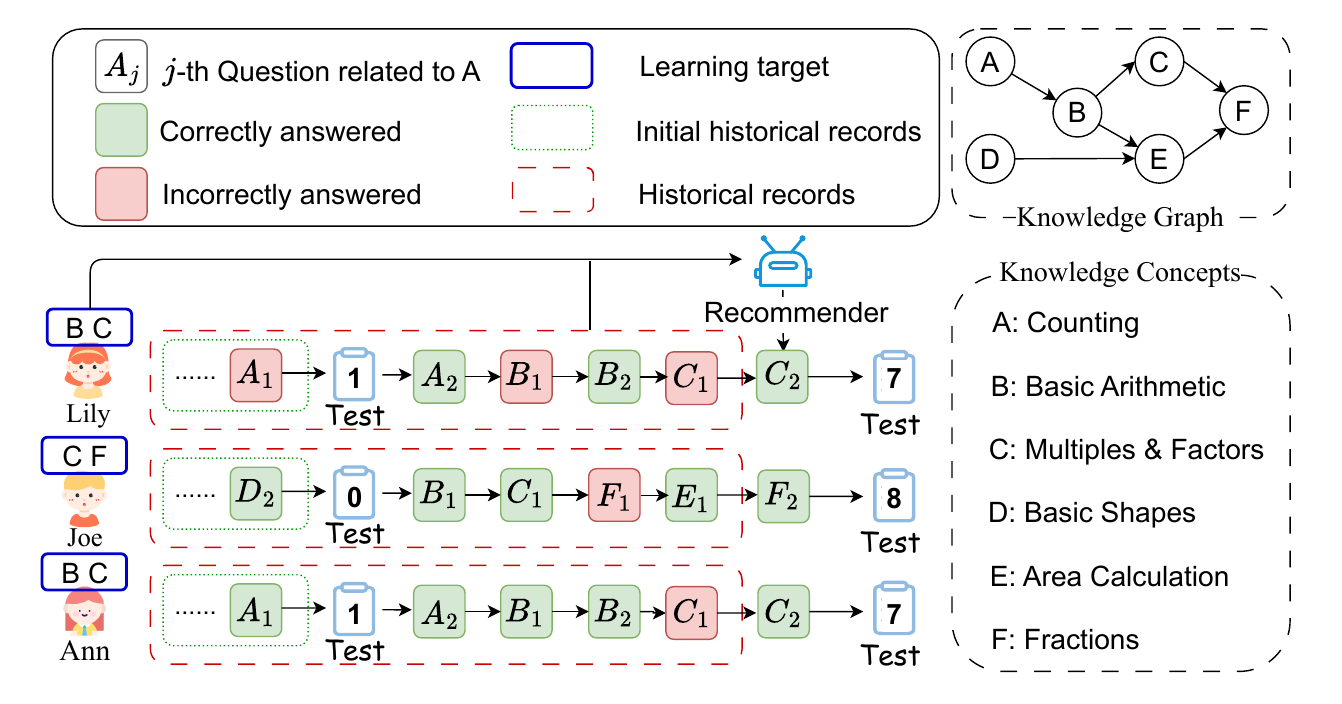} 
\caption{Illustration of the question recommendation task. The knowledge graph indicates the prerequisite relationships between different knowledge concepts.}

\label{fig: intro}

\end{figure}

Question recommendation provides personalized educational services by recommending different questions to help students improve their mastery of knowledge concepts \cite{liu2019exploiting,chen2023set,li2023graph}. By accessing a student's historical records and learning targets (i.e., the specific knowledge concepts they aim to master), the recommender selects a question from a predefined question set to aid in mastering these targets. After the student answers, the recommender updates the historical record and proceeds with the next question. For example, as shown in Figure \ref{fig: intro}, Lily aims to learn \textit{Basic Arithmetic} and \textit{Multiples \& Factors}. Upon realizing that Lily had not yet mastered \textit{Counting}, the recommender initially recommended questions on \textit{Counting} to help her understand foundational concepts. Then, it recommended questions on \textit{Basic Arithmetic} and \textit{Multiples \& Factors} to further support her in mastering her learning target.

To build an efficient recommender and provide personalized educational services, most previous methods \cite{liu2019exploiting,chen2023set,li2023graph} model question recommendation as a Markov Decision Process (MDP) and use reinforcement learning (RL) to maximize students' mastery of learning targets. Specifically, at each training iteration, these methods sequentially recommend questions, collect student feedback, and use this real-time data to optimize the recommendation strategy. By repeatedly performing \textit{recommend questions to collect training data - train the recommender} until the model converges, the recommender can then be used to provide personalized questions for students.

To train a recommender effectively, it is essential to encourage exploration of how different questions impact students' knowledge mastery. Previous studies, such as \cite{liu2019exploiting}, have used classical reinforcement learning exploration strategies like the $\epsilon$-greedy algorithm, where the recommender randomly selects a question with probability $\epsilon$. These strategies focus on individual students and perform well with small student bases and question sets. However, on large-scale platforms with millions of users and questions, such exploration methods struggle to identify suitable questions within limited training iterations, reducing the performance of the recommender.

To address this issue, we propose incorporating collaborative information into the exploration mechanism to improve efficiency in RL-based recommenders. The collaborative principle suggests that similar users should receive similar recommendations, while different users should get distinct ones. As shown in Figure \ref{fig: intro}, Lily and Ann, with the same learning targets, should receive the same questions, while Lily and Joe, with different targets, should receive different questions.

Based on this idea, we introduce Ranking Alignment Recommendation (RAR), a novel exploration mechanism that measure differences between students and differences between recommendations, and align the rankings of these differences. This approach encourages exploration with differentiated recommendations for different students while reduces exploration complexity for similar students. 
RAR consists of two modules: one is the recommendation module, which can be an arbitrary RL-based recommender responsible for recommending questions to students; the other is the ranking alignment module, which helps the recommendation module explore efficiently by calculating and aligning the differences between students and the differences between recommendations.

In summary, the contributions of our paper are:
\begin{itemize}[leftmargin=15pt]
 
    \item We propose a framework named Ranking Alignment Recommender (RAR), designed to help any RL-based question recommender achieve efficient exploration and enhance performance.
    
   \item To our knowledge, we are the first to propose an exploration mechanism based on collaborative information. This new exploration mechanism is expected to benefit any RL-based question recommenders.
   
   \item Our extensive experiments across five settings demonstrate the outstanding performance of our method.
\end{itemize}

\section{Preliminaries} 
\label{preliminaries}

\subsection{Terminologies}
\label{sec:terminologies}
Suppose there is a student $u (u \in \mathcal{U})$ who is learning on the online education platform. We define their \textbf{historical records} at time $t$ as $\mathcal{H}_t^u$ $=\{(q_1, y_1), $ $(q_2, y_2), $ $..., (q_{t-1}, y_{t-1})\}$, where $q_i$ denote the question student $u$ answer at step $i$, and $q_i \in \mathcal{Q}$. $y_i$ denotes the correctness of the student's response to the question $q_i$. The \textbf{learning target} of student $u$ is the set of questions that the student aims to master, denoted as $\mathcal{T}_u \subseteq \mathcal{Q}$. When the student intends to learn specific concepts, the learning target can also be represented as $\mathcal{T}_u = \bigcup_{i} \mathbb{C}_i$, where $\mathbb{C}_i$ denotes the set of questions related to target concept $i$. Following previous works \cite{liu2019exploiting,chen2023set,li2023graph}, we define the \textbf{learning effect} to evaluate the recommendation performance as:
\begin{equation} \label{eq: learning effect}
{\Delta_u=\frac{m_e-m_b}{m_{sup}-m_b}},
\end{equation}
where $m_e$ and $m_b$ denote the mastery of the learning target at the beginning and end of the recommendation respectively, and $m_{sup}$ denotes the maximum mastery of the learning target.

\subsection{Problem Formulation} \label{sec:formulatiom}
In this paper, we aim to design a recommender that, given the historical records $\mathcal{H}_t^u$ and the learning target $\mathcal{T}_u$ of a student $u$ at step $t$, selects a question $q_t$ from the set of questions $\mathcal{Q}$ to recommend to the student. After the student answers the question $q_t$, the historical records $\mathcal{H}_t^u$ will be updated based on the student's response $y_i$, resulting in $\mathcal{H}^u_{t+1}$, which will be used by the recommender for the next step's recommendation. After $n$ rounds of recommendations, the recommendation session ends, and the student's learning effect $\Delta_u$ will be calculated. Our goal is to maximize the students' learning effects $\Delta_u$.

This problem is typically formalized as a Markov Decision Process (MDP) \cite{liu2019exploiting,li2023graph}, where the state $s_t$ is generated by the student's historical records $\mathcal{H}_t^u$ and learning target $\mathcal{T}_u$. The action $a_t$ is the question recommended by the recommender based on $s_t$. The reward $r_t$ represents the reward for taking action $a_t$ in state $s_t$, which is typically determined based on the learning effect $\Delta_u$.

\section{Methodology} \label{method}
\begin{figure*}[t]
\centering
\includegraphics[width=1.0\columnwidth]{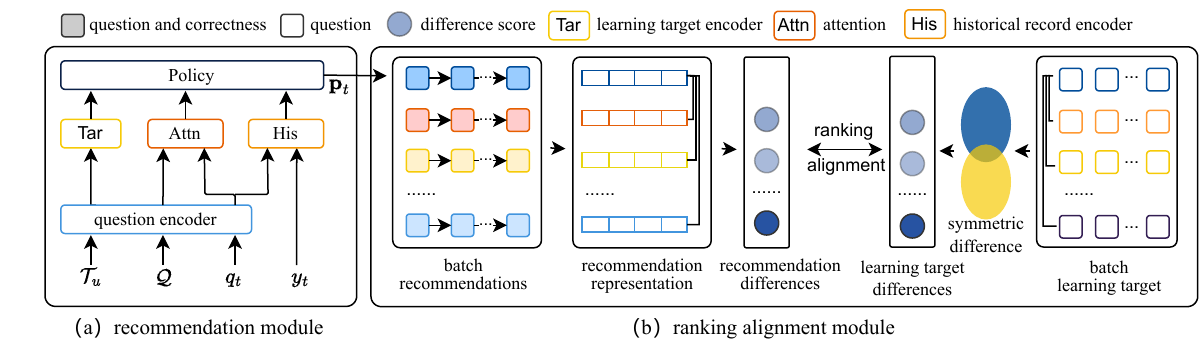}
\caption{The pipeline of our method.}
\label{pipeline}
\end{figure*}

As shown in the Figure \ref{pipeline}, our method includes a recommendation module and a ranking alignment module. The recommendation module is responsible for selecting the recommended questions, while the ranking alignment module is responsible for enhancing the exploration efficiency.

\subsection{Recommendation Module}

The recommendation module receives the students' historical records $\mathcal{H}_{t}^u$ and learning targets $\mathcal{T}_u$ and outputs a recommended question $q_t$ to the students. 

\subsubsection{Encode questions.}
We first encode $q_i$ recommended at time $i$ with one-hot encoding based on the question ID and feed the one-hot vector to an embedding layer. Then, we feed the question embeddings into a multi-head attention layer to capture the relationships between questions, following previous works \cite{lee2019set,chen2023set}:

\begin{equation} \label{eq:ques_emb}
    \bm{e}_i =  f_e(\text{one-hot}(q_i)),
\end{equation}

\begin{equation} \label{eq:ques_rep}
    \bm{r}_i = \textit{Attn}(f_r(\bm{e}_i), \bm{E}, \bm{E}) \oplus f_r(\bm{e}_i),
\end{equation}
where $f_e$, $f_r$ are linear layers.
$e_i \in \mathbb{R}^{d_e}$ denotes the embedding of $q_i$. $E \in \mathbb{R}^{|\mathcal{Q}| \times d_e}$ denote the embedding matrix of all the questions in the question set $\mathcal{Q}$. $\oplus$ denotes concatenation, $\bm{r}_i \in \mathbb{R}^{d_r}$ represents the attentive representation of question $q_i$. \textit{Attn($\cdot, \cdot, \cdot$)} denotes the multi-head attention \cite{vaswani2017attention}, where key and value are $\bm{E}$ and query is $f_r(\bm{e}_i)$.

\subsubsection{Encode Historical records.}
We encode the pair \((q_i, y_i)\) from the students' historical record and input it into a sequence model to obtain a representation of the historical record:

\begin{equation}
    \bm{v}_i=f_v (\bm{e}_{i} \oplus y_i), 
\end{equation}
\begin{equation}
   \bm{h}_{t} = \textit{SEM}(\{\bm{v}_1, \bm{v}_2, ..., \bm{v}_{t-1}\}), 
\end{equation}
where $f_v$ is a linear layer, $\bm{v}_i \in \mathbb{R}^{d_v}$ is the representation vector of $(q_i, y_i)$. \textit{SEM($\cdot$)} denotes a sequential model like Transformer \cite{vaswani2017attention}. $\bm{h}_t \in \mathbb{R}^{d_h}$ denotes the representation of historical record $\mathcal{H}_{t}^u$.

\subsubsection{Encode learning targets.} We use attention representations of the questions $r_i$ included in the learning target to obtain the learning target representation:
\begin{equation} \label{eq:target_mat}
    \bm{a}_t = f_t(\bm{T}), 
    \bm{T} = [\bm{r}_i | q_i \in \mathcal{T}_u],
\end{equation}
where $\bm{T} \in \mathbb{R}^{|\mathcal{T}_u| \times d_r}$ is the target matrix. $f_t$ denotes the learning target encoder, which could implemented by average pooling or multi-head attention \cite{vaswani2017attention}.

\subsubsection{Generate actions.}
Finally, we extract the most recent question from the historical records as a supplementary item, combine the historical record representation and the learning target representation to generate the state $s_t$, and feed it to the policy network to generate the action $a_t$:

\begin{equation} \label{eq: blend}
    \bm{s}_t = f_1(\bm{r}_{t-1}) + f_2(\bm{a}_{t}) + f_3(\bm{h}_{t}),
\end{equation}
\begin{equation} \label{eq:pt}
    \bm{p}_t = \textit{Softmax}(f_p(s_t))
\end{equation}
\begin{equation} \label{eq: sample}
    q_t \sim \bm{p}_t,
\end{equation}
where $f_1$, $f_2$, $f_3$, $f_p$ are linear layers, $\bm{r}_{t-1}$ is the attentive representation of $q_{t-1}$. $\bm{s}_t \in \mathbb{R}^{d_s}$ is the state we encode. $\bm{p}_t \in \mathbb{R}^{|\mathcal{Q}|}$ is the probability of recommending the corresponding question. $q_t \sim \bm{p}_t$ represent sample $q_t$ according to $\bm{p}_t$.

\subsection{Ranking Alignment Module}
\label{RAM}

The Ranking Alignment Module helps the recommender achieve efficient exploration by calculating the differences between students and between recommendations, and aligning their rankings.

\subsubsection{Student differences metrics}
We believe that differences between students are primarily reflected in their learning targets. Therefore, we construct the student differences metric based on the differences in learning targets. Specifically, since a learning target is represented as a set of questions (refer Section \ref{sec:terminologies}), we measure the learning target difference for each pair of students using the cardinality of the symmetric difference: 
\begin{equation} \label{eq:dt}
    \begin{aligned}
    \mathcal{D}_{uv} = \mathcal{T}_u \cup &\mathcal{T}_v - \mathcal{T}_u \cap \mathcal{T}_v \\
    d^t_{uv} &= |\mathcal{D}_{uv}|
    \end{aligned}
\end{equation}
where $\mathcal{T}_u$, $\mathcal{T}_v$ denote the learning target of student $u$, $v$ respectively. $\mathcal{D}_{uv}$ is the symmetric difference of $\mathcal{T}_u$ and $\mathcal{T}_v$. $d^t_{uv}$ denotes the distance of the learning target between $u$ and $v$.

\subsubsection{Recommendation differences metric}  \label{sec: RCD}
We calculate the recommendation difference based on the sequence of questions recommended by the recommender. Specifically, we first obtain the representation of the recommendation sequence, and then compute the recommendation difference based on this representation. 

To make this metric differentiable, we apply the sequence of recommendation probabilities $\bm{p}_t$ in Eq. \ref{eq:pt} to compute the recommendation difference. We tested two methods for representing the recommendation sequence. The first method adopts a sequential model to encode the sequence of recommendation probabilities for an arbitrary student $u$:

\begin{equation} \label{eq:bu1}
    \bm{b}^s_u = f_s(\{\bm{p}_1, \bm{p}_2, ...\}),
\end{equation}
where $f_s(\cdot)$ is a sequential model. $\bm{p}_i$ denotes the probabilities of recommendation at step $i$, which is computed in Eq.\eqref{eq:pt}. $\bm{b}^s_u \in \mathbb{R}^{d_s}$ denotes the \textbf{sequential representation} of the recommendation sequence for student $u$.
We denote the model that encodes the sequence representation using this method as \textbf{RAR-S}.

The second method is to directly sum up the recommended probabilities at each step:
\begin{equation} \label{eq:bu2}
    \bm{b}^a_u = \sum_{i=1}^{n} \bm{p}_i,
\end{equation}
where $n$ is the length of recommendation path, and $\bm{b}^a_u \in \mathbb{R}^{|\mathcal{Q}|}$ denotes the \textbf{addition representation} of the recommendation sequence for student $u$. We denote the model that encodes the sequence representation using this method as \textbf{RAR-A}.

Then, for a pair of students $u$ and $v$, we feed the representation of recommendation sequence to compute the recommendation difference between two students by:
\begin{equation} \label{eq:dp}
    d^p_{uv} = \textit{dist}(\bm{b}_u, \bm{b}_v), 
\end{equation}
where $\bm{b}_u$ can be $\bm{b}_u^s$ or $\bm{b}_u^a$, and $\bm{b}_u$ and $\bm{b}_v$ come from the same encoding method. $\textit{dist}(\cdot)$ denotes the distance between vectors, we implement it with L2 distance. $d^p_{uv}$ denotes the recommendation sequence distance between $u$ and $v$.

\subsubsection{Ranking Alignment}
In each training iteration, for each student $u$ in the batch, we select $m$ students from the current batch to form a student set $\mathcal{U}_u$. 
Then, we apply the user differences $d^t_{uv}$ and the recommendation difference $d^p_{uv}$ between student $u$ and $v \in \mathcal{U}_u$ to construct the rank loss:
\begin{equation} \label{eq:RAC}
    \mathcal{L}_r = \sum_{u \in \mathcal{U}_b} \sum_{v \in \mathcal{U}_u} \textit{Clip}(\psi d^t_{uv} - d^p_{uv}, 0, \omega),
\end{equation}
where $\textit{Clip}(x, a, b)$ is the function that clips the value of $x$ within $a$ and $b$. $\psi$ and $\omega$ are hyperparameter. 

\subsection{Optimization}
We help the recommendation module achieve efficient exploration by adding the rank loss from Eq. \ref{eq:RAC} to the total loss function. For our recommendation module, it will be optimized by three losses:

The first loss is policy gradient loss \cite{silver2014deterministic}:
\begin{equation} \label{eq:pg}
    \mathcal{L}_p = -\sum^T_{t=1} \hat{r}(s_t, q_t) \log {p_t},
\end{equation}
where $p_t$ is the probability of recommending question $q_t$, which is obtained from $\bm{p}_t$ in Eq.\eqref{eq:pt}. $\hat{r}(\bm{s}_t, q_t)$ denotes the accumulative reward of recommendation at step t, which is computed by $\hat{r}(\bm{s}_t, q_t) = r(\bm{s}_t, q_t) + \gamma \hat{r}(\bm{s}_t, q_t)$, $r(\bm{s}_t, q_t)$ is the instant reward of recommendation (refer section \ref{sec:formulatiom}). 

Then, following \cite{chen2023set}, we integrate an auxiliary KT module into the historical record encoder and construct a prediction loss:
\begin{equation}  \label{Loss_k}
\begin{gathered}
    \hat{y}_{t} = f_k(\bm{h}_t), \\
    \mathcal{L}_k = -\sum^{T}_{t=1} (y_{t} \ln{\hat{y}_{t}} + (1-y_{t}) \ln{(1-\hat{y}_{t})}),
\end{gathered}
\end{equation}
where $f_k$ is the KT module, which is implemented by a linear function, $\hat{y}_t$ denotes the corresponding predicted correctness of the question $q_t$.

Finally, with the rank loss in Eq. (\ref{eq:RAC}), the final loss of the recommender is:

\begin{equation} \label{eq:total loss}
    \mathcal{L} = \mathcal{L}_p + \alpha \mathcal{L}_k + \beta \mathcal{L}_r, 
\end{equation}
where $\alpha$ and $\beta$ are hyperparameters.

\section{Experiment}

\subsection{Simulators and datasets} \label{sec: simulator}
Training and evaluating a question recommender relies on students' learning effects, which require interaction with students. However, obtaining these effects is time-consuming, and low-quality recommendations before model convergence may harm students' learning, raising ethical concerns. Therefore, building on previous works \cite{liu2019exploiting,kubotani2021rltutor,chen2023set}, we evaluate our method in simulated environments.

We divide the simulation environments we use into two groups. The first group is constructed using manually designed rules, which includes the KSS \cite{liu2019exploiting} simulator. The second group consists of simulators based on deep knowledge tracing (KT) models. We use two real-world datasets, ASSIST09 \footnote{https://sites.google.com/site/assistmentsdata/home/2009-2010-assistment-data} \cite{feng2009addressing} and Junyi \footnote{https://www.kaggle.com/datasets/junyiacademy/learning-activity-public-dataset-by-junyi-academy} \cite{chang2015modeling}, to train two deep KT models, DKT \cite{piech2015deep} and IEKT \cite{long2021tracing}, respectively, resulting in four deep KT simulators: DKTA09, IEKTA09, DKTJU, and IEKTJU.

\subsection{Baselines} \label{sec: baseline}
We compare our method with three groups of methods.
The first group consists of traditional commercial recommendation methods, including FMLP \cite{zhou2022filter} and GPT-3.5 \cite{brown2020language}. The second group consists of classic reinforcement learning algorithms, including DQN \cite{mnih2013playing} and SAC \cite{haarnoja2018soft}. The third group is recent question recommendation methods, including CSEAL \cite{liu2019exploiting}, SRC \cite{chen2023set}, GEHRL \cite{li2023graph}.

\subsection{Experiment Setting} 

For our proposed method, we set $d_e=48$, $d_r=128$, $d_h=128$, $d_v=128$, $d_s=128$. The head of attention is one for all the cases.
We select the $\alpha$, $\beta$ from $\{0, 0.1, 0.5, 1\}$. We use the Adam \cite{kingma2014adam} to optimize our model. The learning rate is selected from $\{1\times10^{-3}, 5\times10^{-4}, 1\times10^{-4}\}$. We use the learning effect in Eq. \ref{eq: learning effect} as the evaluation metric for the performance of different models.

\renewcommand\arraystretch{1.3}

\begin{table*}[t]
    \centering

    \caption{Comparison of learning effects $\Delta_u$ of different question recommenders in five environments when t=10, 30 and 200.
    Bold indicates the best
    performance among all methods. Underline indicates the second-best performance.}

    \label{tab:effect}
    \scriptsize

    \setlength{\tabcolsep}{0.2mm} {
    \begin{tabular}{c|c c||c c c|c c c|c c c|c c c}
     \hline

        & \multicolumn{2}{c||}{KSS} & 
        \multicolumn{3}{c|}{IEKTJU} & 
        \multicolumn{3}{c|}{DKTJU} & 
        \multicolumn{3}{c|}{IEKTA09} & 
        \multicolumn{3}{c}{DKTA09}   \\
            \cline{1-15}

        $t$ & $10$ & $30$ & 
        $10$ & $30$ & $200$ &
        $10$ & $30$ & $200$ &
        $10$ & $30$ & $200$ &
        $10$ & $30$ & $200$ \\
            \hline \hline

        FMLP & 0.002 & 0.002 &
        -0.053 & -0.074 & -0.087 &
        -0.002 & -0.003 & -0.003 &
        -0.008 & -0.050 & -0.103 &
        -0.054 & -0.073 & -0.074 \\
            \hline

        GPT-3.5 & 0.022 & 0.057 &
        -0.109 & -0.044 & -0.102 &
        -0.003 & -0.001 & -0.000 &
        -0.065 & 0.093 & 0.517 & 
        0.007 & -0.007 & -0.011 \\
            \hline\hline

        DQN & 0.062 & 0.113 & 
        0.169 & 0.293 & 0.525 &
        0.004 & 0.010 & 0.007 &
        0.241 & 0.424 & 0.645 &
        0.019 & 0.016 & 0.005  \\
            \hline

        SAC & 0.031 & 0.049 & 
        -0.020 & 0.007 & 0.137 &
        0.002 & 0.001 & -0.001 &
        0.238 & 0.303 & 0.517 &
        0.011 & 0.010 & 0.020  \\
            \hline

        CSEAL & 0.148 & 0.476 & 
        -0.014 & 0.019 & 0.256 &
        -0.000 & -0.000 & 0.002 &
        -0.029 & -0.070 & 0.099 &
        -0.038 & -0.041 & -0.009  \\
            \hline

        SRC & 0.110 & 0.251 & 
        0.036 & 0.048 & 0.185 &
        0.001 & 0.001 & 0.003 &
        0.073 & 0.132 & 0.311 &
        -0.001 & 0.000 & 0.002  \\
            \hline

        GEHRL & 0.149 & 0.323 &
        0.020 & 0.058 & 0.233 &
        -0.000 & 0.001 & 0.005 &
        0.030 & 0.121 &  0.313 &
        -0.006 & 0.010 & 0.004 \\
            \hline\hline

        RAR-S & \textbf{0.285} & \textbf{0.616} &
        
        \underline{0.265} & \textbf{0.505} & \textbf{0.716} &
        
        \textbf{0.019} & \textbf{0.019} & \textbf{0.019} &
        
        \textbf{0.516} & \textbf{0.576} & \underline{0.677} &
        
        \underline{0.058} & \underline{0.058} & \underline{0.055} \\
            \hline

        RAR-A & \underline{0.232} & \underline{0.614} &
        
        \textbf{0.283} & \underline{0.442} & \underline{0.702} &
        
        \underline{0.013} & \underline{0.012} & \underline{0.016} &
        
        \underline{0.384} & \underline{0.523} & \textbf{0.712} &
        
        \textbf{0.058} & \textbf{0.059} & \textbf{0.059} \\
            \hline
        
	\end{tabular}  } 

\end{table*}

\subsection{Experiment Result}
We measure the performance of different methods after recommending $t$ steps, where $t \in [10, 30, 200]$. Since the maximum steps of recommendation in KSS is 30, we only test the learning effect in $t = 10$ and $t = 30$ for KSS. The results are presented in Table \ref{tab:effect}.
From Table \ref{tab:effect}, we can observe: 
(1) RAR-S and RAR-A achieves the best or second-best performance in all the cases. That demonstrates the effectiveness of our method. 
(2) The second and third groups of baselines, which are based on reinforcement learning, perform better. This indicates that the question recommendation task is more suited to RL-based recommenders, and therefore, our method can benefit most question recommenders.

\subsection{Compatibility Study}

To demonstrate the compatibility of our rank alignment module with different recommendation models, we integrated the module into DQN and SAC and compared their performance. The results, shown in Table \ref{tab:+RAC}, indicate that the rank alignment module improves performance in most cases. For the few instances where performance decreased with DQN, we suspect this is because DQN is unable to effectively learn suitable questions for students from the exploration data.

\renewcommand\arraystretch{1.3}
\begin{table}[t]
    \centering
    \caption{The performance of introducing ranking alignment modules to baselines.}
    \label{tab:+RAC}
    \scriptsize
    \begin{tabular}{c|c|c|c|c|c}
     \hline
        & KSS & DKTA09 & IEKTA09 & DKTJu & IEKTJu \\
        \hline

        DQN & 0.1128 & 0.0050 
        & \textbf{0.6445} & \textbf{0.0070} & 0.5248\\
        \hline

        DQN+RAM & \textbf{0.1310} & \textbf{0.0351} & 0.3736 & 0.0057 & \textbf{0.5463} \\
        \hline \hline

        SAC & 0.0485 & 0.0204 
        & 0.5170 & -0.0011 & 0.1370\\
        \hline

        SAC+RAM & \textbf{0.0799} & \textbf{0.0691} & \textbf{0.7574} & \textbf{0.0085} & \textbf{0.4217} \\
        \hline \hline
        
    \end{tabular}
\end{table}

\section{Conclusion}
In this paper, we propose a framework called Ranking Alignment Recommendation (RAR), which leverages collaborative ideas to effectively enhance the exploration efficiency of RL-based question recommenders. The RAR framework consists of a recommendation module and a rank alignment module, the latter of which can be integrated into any RL-based recommendation module. Through extensive experiments, we demonstrate the effectiveness and compatibility of our framework and highlight the importance of exploration ability for the performance of RL-based models.

\section*{Acknowledgement}
This work is partially supported by the National Natural Science Foundation of China (62307020, U2341229) and the Ministry of Science and Technology of China (2023YFF0905400).

\bibliographystyle{splncs04}
\bibliography{ref}

\end{document}